# Predicting Electricity Consumption using Deep Recurrent Neural Networks

Anupiya Nugaliyadde, Upeka Somaratne, and Kok Wai Wong

Murdoch University, Perth, Australia,
{a.nugaliyadde, upeka.somaratne, k.wong}@murdoch.edu.au

**Abstract.** Electricity consumption has increased exponentially during the past few decades. This increase is heavily burdening the electricity distributors. Therefore, predicting the future demand for electricity consumption will provide an upper hand to the electricity distributor. Predicting electricity consumption requires many parameters. The paper presents two approaches with one using a Recurrent Neural Network (RNN) and another one using a Long Short Term Memory (LSTM) network, which only considers the previous electricity consumption to predict the future electricity consumption. These models were tested on the publicly available London smart meter dataset. To assess the applicability of the RNN and the LSTM network to predict the electricity consumption, they were tested to predict for an individual house and a block of houses for a given time period. The predictions were done for daily, trimester and 13 months, which covers short term, mid-term and long term prediction. Both the RNN and the LSTM network have achieved an average Root Mean Square error of 0.1.

## 1 Introduction

The rapid increase in electricity consumption requires an accurate forecasting of electricity consumption distribution [1]. In order to accurately forecast the electricity usage, the electricity consumption needed to be tracked. Therefore, Advanced Metering Infrastructure (AMI), was introduced. AMI leads to a large amount of electricity consumption data [2]. AMI data is used for electricity consumption forecasting. Forecasting helps make decisions on power distribution from the national grid. An accurate forecast on the electricity consumption can prevent unplanned electricity distribution disruptions [1, 3-5]. AMI provides the background to utilize data for descriptive, predictive and prescriptive analytics [2]. The demand for energy is based on various factors such as weather, occupancy, types of machines and appliances used. The dependency on high number of factors have made forecasting techniques much complex. Accurate predictions of the electricity consumption is important for efficient distribution. However, applying all the variables that effect electricity consumption can create a complex forecasting model which is unstable and unpredictable [2]. Therefore, data-driven solutions to predict electricity consumption focuses on time-series solutions [3].

Electricity consumption is a time-dependent attribute. Therefore, there are approaches that use time series to build the model to predict electricity consumption. Availability of past information leads to solutions based on time series analysis since it reflects the time-dependent variations [6]. The forecasts for electricity consumption have been identified as short term (hourly to one week), mid-term (one week to one year), and long term (more than one year) forecasts [2]. Time-series analysis techniques are addressed using conventional approaches and AI-based approaches (ANN, ARIMA, SVM, Fuzzy based techniques) [3, 5]. Past research shows these techniques perform better for short term forecasting but poor in mid-term and long term forecasting [3]. The research conducted on mid-term to long term forecasting shows an excess of 40%-50% in relative errors [1]. There are many challenges for mid-term and long term electricity consumption forecasting [3], and thus form the focus of this paper.

This paper presents two approaches, a RNN and a LSTM, to forecast electricity consumption for short-term, mid-term and long-term. The RNN and the LSTM were used to predict daily, trimester and thirteen monthly electricity consumption. The RNN and the LSTM are compared with the most common and popular electricity consumption prediction models (ARIMA, ANN and DNN). Both models have shown to minimize the root mean square error compared to the other models. The models were tested on the publicly available London Smart Meter dataset. The experiments were conducted on predicting both an individual houses electricity consumption and a block of houses electricity consumption. The LSTM and RNN have achieved, on average a Root Mean Square Error (RMSE) of 0.1 for all cases.

## 2    Related Work

Smart grid data has been used for many electricity consumption forecasting tasks [2]. The data is treated as sequential data. Most commonly Autoregressive integrated moving average (ARIMA) [7], Support Vector Machines (SVM) [8], linear regression [9], and Artificial Neural Networks (ANN) [10] were tested in order to forecast the electricity consumption.

ARIMA is one of the most common technique used for time series forecasting. ARIMA models applied to forecast household consumption and predict demand for office buildings [11]. Furthermore, ARIMA was used for short-term forecasting for half-hourly consumption in Malaysia [12]. The results of models have shown high performances for short term predictions [11]. ARIMA models have been identified as a high performing solution for short term predictions [3]. SVM has also been used for forecasting electricity consumption [8]. However, the non-linear models have shown to achieve better results for short term prediction [10]. A comparison with ANN, Multiple Regression (MR), Genetic Programming (GP), DNN and SVM, the results showed higher performance for ANN. The experiment results showed that despite that the amount of data was limited, the DNN has produced comparable results with other techniques tested in the research [4].

A model which require no/less feature selections could immensely improve a models predictions [6]. Therefore, deep learning models have a higher capability of accurately predicting on time series data [6] [13] [14]. DNNs which use past inputs to predict the future have demonstrated the capability in achieving more accurate results in sequential data [15]. However, these learning models have underperformed in mid-term prediction and long-term prediction.

RNN and LSTM are the most prominent deep neural networks which have a feedback loop from the past inputs [16]. The LSTM and RNN have shown to outperform other feed forward deep learning techniques (DNN) which do not hold feedback loops [17]. There are complex versions of the deep learning models with feedback loops, however, before moving on to complex models [18], which are high in computational cost. It is effective to test on the basic models before moving on to complex deep learning models. A self-recurrent wavelet neural network is proposed to generate accurate results on load forecasting [2]. However, it focus on short term electricity consumption forecasting only. A pooling based deep RNN was introduced to learn spatial information and address the overfitting issues that other methods had [19]. This has shown that it outperforms the SVM and ARIMA for the given tasks. However, requiring many models to predict electricity consumption for all terms (short term, mid-term and long term) is a tedious tasks. Therefore, a requirements of having one model which is capable of predicting electricity consumption is needed.

## 3 Methodology[1]

Deep learning is capable of learning from hidden patterns with no feature selection and outperform most of the machine learning and statistical methods to achieve various tasks [20, 21] [22] [23] [24]. Time series data holds a sequential pattern, in which the data holds co-relationships between parallel data instances ($x_t$ depends on $x_{t-1}$ and $x_t$ effects $x_{t+1}$). Sequential data is handled by Recurrent Neural Networks (RNN), Long Short-Term Memory Networks (LSTM) and memory networks due to the capability of memory to hold past information.

### 3.1 Data pre-processing

In order to predict an individual houses, electricity consumption data is not preprocessed because it is important to focus on models that relies less on preprocessing. The data for an individual house is separated from the blocks' electricity consumption. In order to predict a blocks electricity consumption, the data had to be pre-processed as follows. A block electricity consumption is predicted using all the houses electricity consumption per day and the mean of the electricity consumption per day.

The time periods of each house is not consistent throughout a given block. Therefore, a common time period which most of the houses are involved in a given block is taken for block predictions. The mean electricity consumption for a given day is calculated for the above selected time interval allowing a larger set of houses to calculate the mean value. An example mean value calculated for the block 36 is given[2]. The mean value would generate one value to a given date for the block.

---

[1] The code can be found at https://github.com/anupiyan/Time-series-Recurrent-Neural-Network

Predictions were performed using RNN and LSTM the most common deep time series, prediction models.

### 3.2    Setting up training and testing data

The dataset is divided into training and testing data. The testing data is kept separate from the training data. Therefore, the testing data is unseen to the model until testing the models. The models were trained on 80% of all data and tested on 20% of all the data.

The training and testing data is divided into three different methods. In order to understand the electricity consumption of an individual house and a block of houses, the training data is set up.

1. Given one day predicting the next day
   This is done for the blocks of houses and individual houses. The model will be trained to take one day's input and predict the next day's electricity consumption.
2. Given one month and predicting the next month. In order to get a balanced dataset, only 600 days are considered. This method is applied to individual houses and blocks mean value predictions.
3. Given three months and predicting the next three months. Balance dataset is achieved by using 600 days. Three months were given to predict three months ahead.

However, in order to show that the models are capable of learning from a short term, mid-term and long term, the data is divided. The short term predictions are shown by predicting one day ahead, the mid-term prediction is shown by predicting 3 months ahead, and the long term is shown by predicting 13 months ahead. The input $x_t$ is used to predict the output ($x_{t+1}$). $x_t$ can be one day's, three months and 13 months electricity consumption for an individual house or block of houses.

### 3.3    RNN

RNN is deep learning model which learn from sequences [6]. RNN recursively takes the past output ($y_{t-1}$) and adds it to the current input ($x_t$). RNN current output ($y_t$) learns from the past sequence, using the $y_{t-1}$. Fig.1 shows one RNN unit structure. The sequence is passed on as an input to the current input. Therefore, $y_t$ would be influenced by the past sequences. The past sequences carries on the past results with a combination. Therefore, sequential information effects all the outputs and carried on throughout a given sequence. (1) presented the RNN's equation to demonstrate the recursion of the $y_{t-1}$ to the $x_t$. $W_r$ is the weight given to $y_{t-1}$ and $W_n$ is the weight given to the $x_t$. In our experiment, the RNN has 100 hidden units connected to each other sequentially to generate the best results.

---

[2]    https://github.com/anupiyan/Time-series-Recurrent-Neural-Network/blob/master/block%2036%20Mean%20value%20per%20day.csv

$$y_t = tanh(W_r y_{t-1} + W_n x_t) \qquad (1)$$

The RNN is trained for 300 epochs using a 20 batch size. Adam optimizer is used for the optimization to generate the highest accuracy. Fig. 3 shows the RNN structure applied in the experiment.

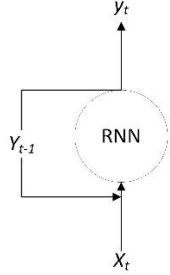

**Fig. 1.** The RNN architecture. $x_t$ is the current input, $y_{t-1}$ is the previous output which is passed and combined with $x_t$ to create the final output $y_t$.

### 3.4 LSTM

The LSTM has a complex architecture compared with the RNN. Fig. 2 shows an LSTM cell architecture, which includes three gates that are used to filter and carry forward the past data [17]. The input $x_t$ is passed and added to the $h_{t-1}$. The $f_t$ gate decides if $C_{t-1}$ is carried forward to generate the current output ($h_t$). Unlike the RNN which carries on $y_{t-1}$, the LSTM decides on which data is carried forward through the cell state ($C_{t-1}$). Furthermore, unlike RNN, the LSTM cell contains the input gate and the output gate to create the final output. $C$ carries the $h_{t-1}$ to the next time stamp. However, unlike RNN the LSTM calculates the $C$. LSTM therefore, has a controlled over the $C$ and the $h_t$ rather than directly generating the output. LSTM cell is considered as a unit, and the unit can be sequentially connected to each other. The LSTM has 100 LSTM units and the last layer has one dense layer before generating the final output. The LSTM used in the experiment ran for 300 epochs to train to the optimal results. The Adam optimizer was used to generate the highest accurate results. 20 batch size was used for the LSTM. Fig. 3 shows how the LSTM structure applied in the experiment.

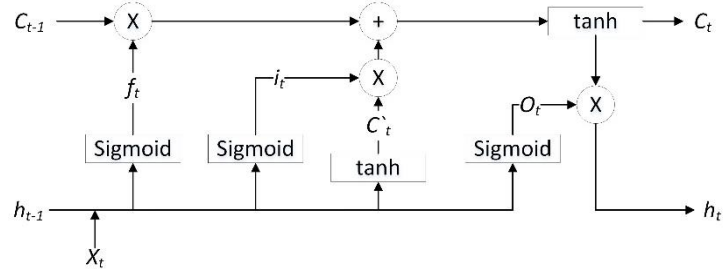

**Fig. 2.** Long Short Term Memory Network (LSTM) cell architecture

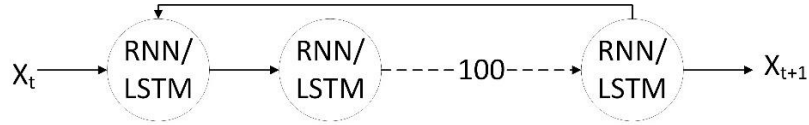

**Fig. 3.** Is the structure of the LSTM or RNN. The model has 100 hidden layers. The model takes input and predicts batch. $x_t$ and $x_{t+1}$ can be a day, month or 3 months.

## 4    Results and Discussion

### 4.1    Data

The experimentation is conducted using the publicly available London Smart Meter data set[3]. This dataset has a refactored version of the London data stores. The dataset comprises of real-world 5567-meter data of UK households electricity consumption. The data set also holds weather for the given dates and the information (number of occupants, number of rooms, and etc.) of each household. The experimentation only focuses on the electricity consumption. The data was collected from November 2011 to February 2014. The meter data is divided into blocks, depending on the location of the house. Each block contains around 15-20 houses. Data can be used to predict an individual houses electricity consumption or block of houses.

### 4.2    Results

The results are divided into three categories, based on the prediction time, either short term, mid-term and long term predictions. Furthermore, the predictions are made for an individual house, a block of houses, and the median value of a block of houses per day.
The experiments were conducted to predict two levels of categories. They are predicting the electricity consumption for a given household and predicting a blocks of electricity consumption. The predictions are evaluated using the Root Mean Square Error (RMSE). The results of RNN and LSTM are compared with ARIMA, Artificial Neural Networks (ANN) and Deep Neural Networks (DNN).

### 4.3    Short Term Predictions

In order to demonstrate short term predictions, the models are used to predict electricity consumption one day in advance for individual house and a block of houses.

**Predicting electricity consumption for a given household**

---

[3] https://www.kaggle.com/jeanmidev/smart-meters-in-london

The randomly selected set of individual houses are used to demonstrate the performance of each models. The results are shown in Table 1. In comparison to RNN, LSTM, ANN and DNN, ARIMA has shown to perform better for short term predictions. However, LSTM and RNN have also shown very close performances to ARIMA.

**Table 1.** Short term predictions (one day ahead). The results are in RMSE.

| House number | ARIMA | ANN | DNN | RNN | LSTM |
|---|---|---|---|---|---|
| MAC000002 | 0.06 | 0.2 | 0.13 | 0.08 | 0.08 |
| MAC000033 | 0.02 | 0.25 | 0.17 | 0.03 | 0.03 |
| MAC000092 | 0.03 | 0.26 | 0.16 | 0.03 | 0.03 |
| MAC000156 | 0.01 | 0.27 | 0.19 | 0.03 | 0.04 |
| MAC000246 | 0.11 | 0.29 | 0.21 | 0.12 | 0.12 |
| MAC00450  | 0.2 | 0.3 | 0.23 | 0.19 | 0.2 |
| MAC001074 | 0.1 | 0.24 | 0.19 | 0.11 | 0.12 |
| MAC003223 | 0.07 | 0.3 | 0.2 | 0.09 | 0.09 |

**Predicting electricity consumption using all the houses in a block data**
The first approach to predict electricity consumption was to consider all the houses electricity consumption. All the values from each house in the block are used to predict the electricity consumption for the whole block. The results are shown in Table 2.

**Table 2.** Short term predictions for a selected set of blocks. The results are in RMSE.

| Block | ARIMA | ANN | DNN | RNN | LSTM |
|---|---|---|---|---|---|
| 0 | 0.4 | 0.55 | 0.5 | 0.47 | 0.45 |
| 25 | 0.2 | 0.34 | 0.36 | 0.25 | 0.25 |
| 36 | 0.15 | 0.29 | 0.28 | 0.22 | 0.22 |
| 51 | 0.07 | 0.12 | 0.13 | 0.07 | 0.07 |
| 61 | 0.05 | 0.1 | 0.09 | 0.07 | 0.08 |
| 90 | 0.03 | 0.13 | 0.11 | 0.05 | 0.06 |
| 108 | 0.04 | 0.11 | 0.16 | 0.09 | 0.08 |

**Predicting Electricity Consumption for a given block of houses using the mean consumption per day**
As observed in Table 2, the results from the RNN and the LSTM have not performed comparatively to the individual houses prediction result. In order to achieve better results in predicting the electricity consumption for a given block of houses, the mean consumption per day was calculated. Table 3 shows the results of predicting each block mean consumption per day. Table 3 shows a vast improvement over Table 2, which gives a clear insight into the electricity consumption per day in a block of houses.

Table 3. Short Term prediction on the mean value for each day for each block. The results are in RMSE.

| Block | ARIMA | ANN | DNN | RNN | LSTM |
|---|---|---|---|---|---|
| 0 | 0.03 | 0.13 | 0.11 | 0.03 | 0.03 |
| 25 | 0.01 | 0.11 | 0.1 | 0.02 | 0.01 |
| 36 | 0.01 | 0.1 | 0.09 | 0.01 | 0.02 |

### 4.4   Mid-term Predictions

In order to show the capability of mid-term predictions, models are established using data from past three months' electricity consumption to predict the next three months. To be consistent in comparing the short term, mid-term and long term predictions, the same set of houses and block are used.

**Predicting electricity consumption for a given household**
Table 4 shows that, unlike the short term predictions, in mid-term predictions, ARIMA has not performed as good as other methods. All three deep neural networks (DNN, RNN and LSTM) has shown to outperform ARIMA and ANN. These results show that deep hidden models have outperformed the simpler models.

Table 4. Mid-term predictions (three months ahead). The results are shown in RMSE.

| House Number | ARIMA | ANN | DNN | RNN | LSTM |
|---|---|---|---|---|---|
| MAC000002 | 0.27 | 0.33 | 0.23 | 0.1 | 0.09 |
| MAC000033 | 0.25 | 0.32 | 0.25 | 0.06 | 0.07 |
| MAC000092 | 0.21 | 0.39 | 0.2 | 0.05 | 0.06 |
| MAC000156 | 0.14 | 0.29 | 0.21 | 0.06 | 0.07 |
| MAC000246 | 0.34 | 0.35 | 0.27 | 0.14 | 0.16 |
| MAC00450 | 0.39 | 0.4 | 0.24 | 0.22 | 0.22 |
| MAC001074 | 0.29 | 0.31 | 0.3 | 0.15 | 0.16 |
| MAC003223 | 0.31 | 0.45 | 0.26 | 0.12 | 0.13 |

**Predicting electricity consumption using all the houses in a block data**
To be consistent, similar blocks are used here, which have been used for short term predictions. Table 5 shows that similarly to individual house electricity consumption prediction, the deep learning models have outperformed the rest of the models. However, the LSTM and RNN have outperformed the DNN as well.

Table 5. Mid-term predictions for a selected set of blocks. The results are in RMSE.

| Block | ARIMA | ANN | DNN | RNN | LSTM |
|---|---|---|---|---|---|
| 0 | 0.51 | 0.6 | 0.54 | 0.3 | 0.3 |
| 25 | 0.32 | 0.32 | 0.36 | 0.21 | 0.21 |
| 36 | 0.35 | 0.35 | 0.32 | 0.2 | 0.2 |
| 51 | 0.19 | 0.24 | 0.16 | 0.09 | 0.09 |
| 61 | 0.15 | 0.2 | 0.19 | 0.1 | 0.1 |
| 90 | 0.03 | 0.13 | 0.11 | 0.05 | 0.06 |
| 108 | 0.04 | 0.11 | 0.16 | 0.09 | 0.08 |

**Predicting Electricity Consumption for a given block of houses using the mean consumption per day**

Table 6 shows the mid-term electricity consumption. This also shows that the deep neural networks outperformed the other time series methods. Furthermore, RNN and LSTM have shown to achieve more accurate results. The RNN and LSTM have performed similarly and have a higher marginal difference compared to the other approaches.

Table 6. Mid-Term prediction on the mean value for each day for each block. The results are in RMSE.

| Block | ARIMA | ANN | DNN | RNN | LSTM |
|---|---|---|---|---|---|
| 0 | 0.22 | 0.22 | 0.17 | 0.04 | 0.05 |
| 25 | 0.2 | 0.21 | 0.15 | 0.02 | 0.02 |
| 36 | 0.24 | 0.22 | 0.14 | 0.02 | 0.02 |

### 4.5 Long term Prediction

In the experiments, long term predictions are focused on predicting over 13 months in advance. The models are trained to predict 13 months in advance, given the first month. In this results, as shown in Table 7, 8 and 9. The overall RMSE increases. However, compared to the other models, RNN and LSTM have achieved considerably lower RMSE values.

**Predicting electricity consumption for a given household**

Table 7 showed the long term predictions results for a given household. RNN and LSTM are capable of predicting with more accurate results compared to the other models.

**Table 7.** Long term predictions (thirteen months ahead). A comparison of ARIMA, ANN, DNN, RNN and LSTM is conducted. The results are shown in RMSE.

| House number | ARIMA | ANN | DNN | RNN | LSTM |
|---|---|---|---|---|---|
| MAC000002 | 0.44 | 0.39 | 0.3 | 0.12 | 0.15 |
| MAC000033 | 0.55 | 0.35 | 0.28 | 0.11 | 0.13 |
| MAC000092 | 0.51 | 0.39 | 0.28 | 0.12 | 0.13 |
| MAC000156 | 0.56 | 0.32 | 0.3 | 0.13 | 0.16 |
| MAC000246 | 0.54 | 0.39 | 0.3 | 0.15 | 0.17 |
| MAC00450 | 0.57 | 0.41 | 0.35 | 0.24 | 0.24 |
| MAC001074 | 0.58 | 0.44 | 0.34 | 0.19 | 0.21 |
| MAC003223 | 0.6 | 0.4 | 0.34 | 0.2 | 0.2 |

**Predicting electricity consumption using all the houses in a block data**

Similar to the individual house, the block's electricity consumption is predicted using the same 13 months ahead prediction. The LSTM and RNN have shown to outperform the rest of the models. Furthermore, the results also show that all neural network models have outperformed ARIMA.

**Table 8.** Long term predictions for a selected set of blocks. The results are in RMSE.

| Block | ARIMA | ANN | DNN | RNN | LSTM |
|---|---|---|---|---|---|
| 0 | 0.6 | 0.51 | 0.41 | 0.24 | 0.26 |
| 25 | 0.56 | 0.42 | 0.38 | 0.2 | 0.22 |
| 36 | 0.53 | 0.4 | 0.46 | 0.24 | 0.28 |
| 51 | 0.52 | 0.4 | 0.35 | 0.2 | 0.21 |
| 61 | 0.44 | 0.41 | 0.35 | 0.22 | 0.27 |
| 90 | 0.49 | 0.5 | 0.43 | 0.21 | 0.25 |
| 108 | 0.51 | 0.44 | 0.37 | 0.2 | 0.2 |

**Predicting Electricity Consumption for a given block of houses using the mean consumption per day**

Similar to the other results with long term prediction, RNN and LSTM have shown to outperform all the other models. This has suggested that the capability of handling longer sequence information can be taken care by RNN and LSTM.

**Table 9.** Long Term prediction on the mean value for each day for each block. The results are in RMSE.

| Block | ARIMA | ANN | DNN | RNN | LSTM |
|---|---|---|---|---|---|
| 0 | 0.4 | 0.38 | 0.32 | 0.14 | 0.15 |
| 25 | 0.43 | 0.4 | 0.37 | 0.13 | 0.15 |
| 36 | 0.46 | 0.39 | 0.34 | 0.14 | 0.14 |

## 5  Conclusion

The paper focuses on models which can be used to predict electricity consumption for an individual house and a block of houses for short term, mid-term and long term. The paper compares ARIMA, ANN, DNN, RNN and LSTM by conducting experiments for the publicly available London Smart Meter dataset. Although ARIMA has shown to perform well for short term predictions, it is clear that as the length of time increases, ARIMA does not perform well compared to the other models. RNN and LSTM have shown to perform similar to ARIMA for short term predictions while outperformed all the other models in mid-term and long term predictions. It is evident through the London Smart Meter dataset that the RNN and LSTM is capable of predicting short term, mid-term and long term forecasts for electricity consumption with high accuracy.

## References

1. Rahman, A., Srikumar, V., Smith, A.D.: *Predicting electricity consumption for commercial and residential buildings using deep recurrent neural networks.* Applied energy. **212**, 372-385 (2018).
2. Wang, Y., Chen, Q., Hong, T., Kang, C.: *Review of smart meter data analytics: Applications, methodologies, and challenges.* IEEE Transactions on Smart Grid. (2018).
3. Deb, C., Zhang, F., Yang, J., Lee, S.E., Shah, K.W.: *A review on time series forecasting techniques for building energy consumption.* Renewable and Sustainable Energy Reviews. **74**, 902-924 (2017).
4. Amber, K., Ahmad, R., Aslam, M., Kousar, A., Usman, M., Khan, M.: *Intelligent techniques for forecasting electricity consumption of buildings.* Energy. **157**, 886-893 (2018).
5. Cai, H., Shen, S., Lin, Q., Li, X., Xiao, H.: *Predicting the energy consumption of residential buildings for regional electricity supply-side and demand-side management.* IEEE Access. (2019).
6. Boukoros, S., Nugaliyadde, A., Marnerides, A., Vassilakis, C., Koutsakis, P., Wong, K.W.: *Modeling server workloads for campus email traffic using recurrent neural networks*. In: *International Conference on Neural Information Processing*, pp. 57-66. Springer, (2017).
7. Barak, S.,Sadegh, S.S.: *Forecasting energy consumption using ensemble ARIMA–ANFIS hybrid algorithm.* International Journal of Electrical Power & Energy Systems. **82**, 92-104 (2016).
8. Ahmad, A., Hassan, M., Abdullah, M., Rahman, H., Hussin, F., Abdullah, H., Saidur, R.: *A review on applications of ANN and SVM for building electrical energy consumption forecasting.* Renewable and Sustainable Energy Reviews. **33**, 102-109 (2014).
9. Bianco, V., Manca, O., Nardini, S.: *Electricity consumption forecasting in Italy using linear regression models.* Energy. **34**(9), 1413-1421 (2009).

10. Darbellay, G.A.,Slama, M.: *Forecasting the short-term demand for electricity: Do neural networks stand a better chance?* International Journal of Forecasting. **16**(1), 71-83 (2000).
11. Chou, J.-S.,Tran, D.-S.: *Forecasting energy consumption time series using machine learning techniques based on usage patterns of residential householders.* Energy. **165**, 709-726 (2018).
12. Mohamed, N., Ahmad, M.H., Ismail, Z.: *Short term load forecasting using double seasonal ARIMA model.* In: *Proceedings of the regional conference on statistical sciences*, pp. 57-73. (2010).
13. Tan, Y., Liu, W., Qiu, Q.: *Adaptive power management using reinforcement learning.* In: *Computer-Aided Design-Digest of Technical Papers, 2009. ICCAD 2009. IEEE/ACM International Conference on*, pp. 461-467. IEEE, (2009).
14. Yuan, J., Wang, Y., Wang, K.: *LSTM Based Prediction and Time-Temperature Varying Rate Fusion for Hydropower Plant Anomaly Detection: A Case Study.* In: *International Workshop of Advanced Manufacturing and Automation*, pp. 86-94. Springer, (2018).
15. Hochreiter, S.,Schmidhuber, J.: *Long short-term memory.* Neural computation. **9**(8), 1735-1780 (1997).
16. Nugaliyadde, A., Wong, K.W., Sohel, F., Xie, H.: *Language Modeling through Long Term Memory Network.* arXiv preprint arXiv:1904.08936. (2019).
17. Gers, F.A., Schmidhuber, J., Cummins, F.: *Learning to forget: Continual prediction with LSTM.* (1999).
18. Kumar, A., Irsoy, O., Ondruska, P., Iyyer, M., Bradbury, J., Gulrajani, I., Zhong, V., Paulus, R., Socher, R.: *Ask me anything: Dynamic memory networks for natural language processing.* In: *International Conference on Machine Learning*, pp. 1378-1387. (2016).
19. Shi, H., Xu, M., Li, R.: *Deep learning for household load forecasting—A novel pooling deep RNN.* IEEE Transactions on Smart Grid. **9**(5), 5271-5280 (2017).
20. LeCun, Y., Bengio, Y., Hinton, G.: *Deep learning.* Nature. **521**(7553), 436-444 (2015).
21. Sabour, S., Frosst, N., Hinton, G.E.: *Dynamic routing between capsules.* In: *Advances in Neural Information Processing Systems*, pp. 3859-3869. (2017).
22. Ashangani, K., Wickramasinghe, K., De Silva, D., Gamwara, V., Nugaliyadde, A., Mallawarachchi, Y.: *Semantic video search by automatic video annotation using TensorFlow.* In: *2016 Manufacturing & Industrial Engineering Symposium (MIES)*, pp. 1-4. IEEE, (2016).
23. Nugaliyadde, A., Wong, K.W., Sohel, F., Xie, H.: *Reinforced memory network for question answering.* In: *International Conference on Neural Information Processing*, pp. 482-490. Springer, (2017).
24. Ren, S., He, K., Girshick, R., Sun, J.: *Faster r-cnn: Towards real-time object detection with region proposal networks.* In: *Advances in neural information processing systems*, pp. 91-99. (2015).